\documentclass[11pt]{article}

\usepackage[preprint]{acl}

\usepackage{times}
\usepackage{latexsym}
\usepackage[T1]{fontenc}
\usepackage[utf8]{inputenc}
\usepackage{microtype}
\usepackage{inconsolata}
\usepackage{graphicx}

\usepackage{amsmath,amssymb}
\usepackage{bm}
\usepackage{multirow}
\usepackage{booktabs}
\usepackage[table]{xcolor}
\usepackage{tcolorbox}
\tcbuselibrary{breakable}

\title{
Policy Split: Incentivizing Dual-Mode Exploration in LLM Reinforcement
\\with Dual-Mode Entropy Regularization
}
\author{
    Jiashu Yao\textsuperscript{\rm 1},
    Heyan Huang\textsuperscript{\rm 1},
    Daiqing Wu\textsuperscript{\rm 2}
    Zeming Liu\textsuperscript{\rm 3},
    Yuhang Guo\textsuperscript{\rm 1}\thanks{Corresponding author.}
    \\
    \textsuperscript{\rm 1}Beijing Institute of Technology \
    \textsuperscript{\rm 2}Tsinghua University \
    \textsuperscript{\rm 3}Beihang University \\
}

\begin{document}
\maketitle
\begin{abstract}
To encourage diverse exploration in reinforcement learning (RL) for large language models (LLMs) without compromising accuracy, we propose Policy Split, a novel paradigm that bifurcates the policy into normal and high-entropy modes with a high-entropy prompt. While sharing model parameters, the two modes undergo collaborative dual-mode entropy regularization tailored to distinct objectives. Specifically, the normal mode optimizes for task correctness, while the high-entropy mode incorporates a preference for exploration, and the two modes learn collaboratively. Extensive experiments demonstrate that our approach consistently outperforms established entropy-guided RL baselines across various model sizes in general and creative tasks. Further analysis reveals that Policy Split facilitates dual-mode exploration, where the high-entropy mode generates distinct behavioral patterns to the normal mode, providing unique learning signals. Code available at \url{https://github.com/BITHLP/PolicySplit}.
\end{abstract}

\section{Introduction}

The application of reinforcement learning (RL) on large language models (LLMs) \cite{schulman2017proximal, shao2024deepseekmath, liu2025understanding} has demonstrated significant success, most notably in the emergence of the o1- and R1-like large reasoning models (LRMs) \cite{guo2025deepseek, qwq32b, team2025kimi, seed2025seed1}. These LRMs exhibit sophisticated reasoning capacity, particularly in domains requiring complex problem-solving and mathematical proficiency.

Building on the success of RL for LLMs, recent research has drawn inspirations from entropy-regularized RL \cite{ziebart2008maximum, mnih2016asynchronous, haarnoja2018soft} to develop adapted RL frameworks \cite{cheng2025reasoning, yao2025diversity, tan2025gtpo}, aiming to incentivize high-entropy rollouts for more effective exploration. These methods have proven successful in mitigating entropy collapse during training, ultimately enhancing the final accuracy and reasoning robustness.

\begin{figure}[tb!]
    \centering
    \includegraphics[width=\linewidth]{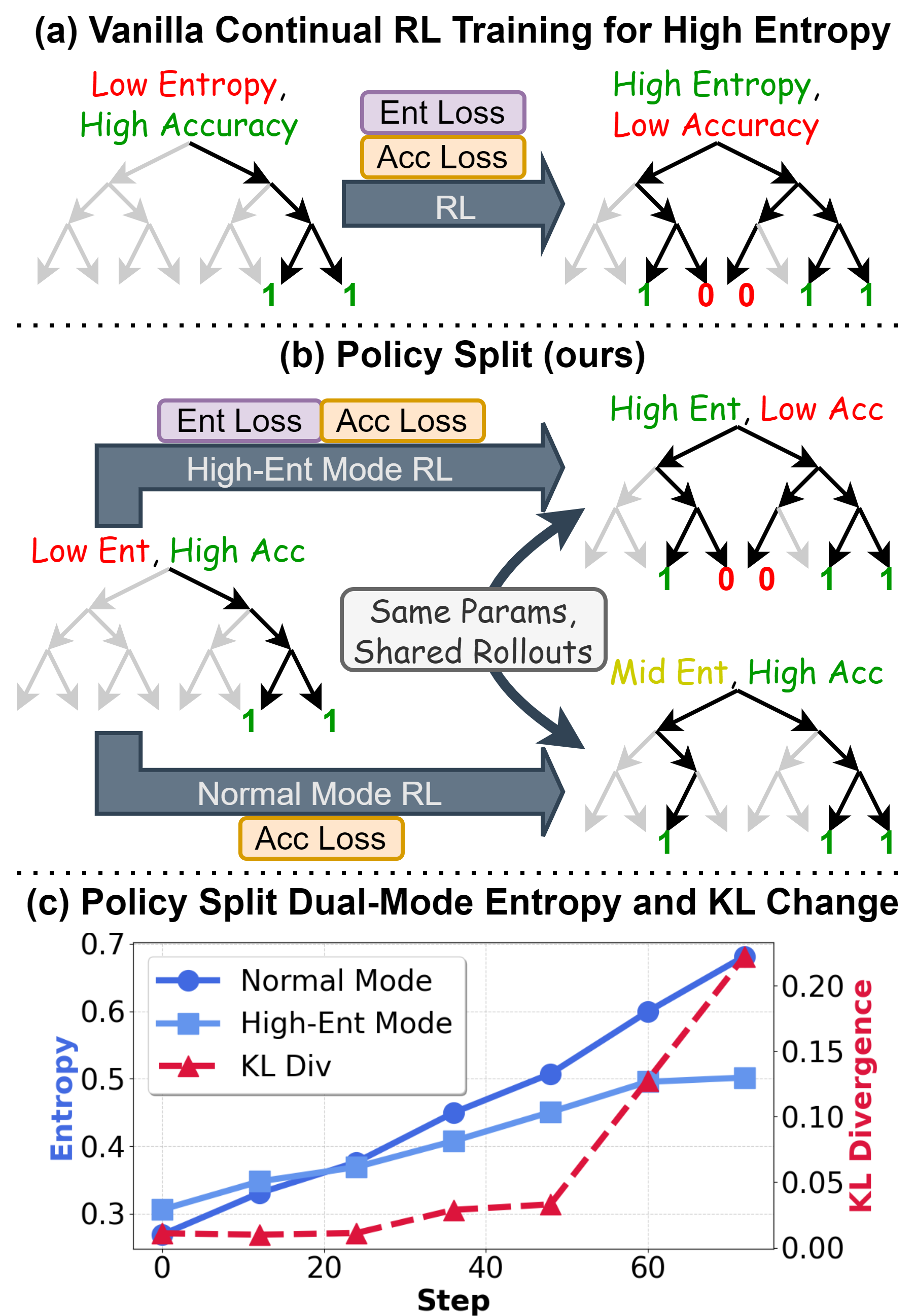}
    \caption{(a) Entropy-guided RL can leads to accuracy drop when prefering high entropy. (b) Unlike previous work, Policy Split innovatively bifurcates the policy into normal and high-entropy modes. (c) Through dual-mode entropy regularization in Policy Split, two modes obtain a larger KL divergence and exhibit distinct entropy behaviors.}
    \label{fig:intro}
\end{figure}

However, a fundamental conflict between exploration and exploitation persists in entropy-guided RL for LLMs. Recent studies suggest that performance gains are often strongly correlated with entropy reduction \cite{yue2025does, cui2025entropy, gao2025one, zhang2025right}, implying that incentivizing higher entropy may inadvertently degrade accuracy. More directly, recent paper \cite{jiang2025rethinking} demonstrates that naive entropy regularization can cause outputs to degenerate into incoherent high-entropy sequences. As illustrated in Figure \ref{fig:intro}(a), this performance degradation is particularly acute during the continual post-training of low-entropy LRMs. Since these models already possess robust capabilities, forcing higher entropy tends to undermine their established proficiency.

To maintain high accuracy while encouraging exploration during the continual post-training of LRMs, we adopt a simple intuition, that we can instantiate two policies within one model. One policy is optimized for accuracy, while the other prioritizes high entropy, and these two policies benefit from each other's experiences.

Specifically, we propose Policy Split, a novel paradigm for entropy-guided LRM reinforcement learning. In Policy Split, the policy is firstly bifurcated into normal mode policy and high-entropy mode policy, where the latter one is identified with an additional high-entropy system prompt. During training, as is shown in Figure \ref{fig:intro}(b), the normal mode is trained towards maximized correctness the same as original GRPO-like RL, while the high entropy mode policy additionally adds a high-entropy preference to encourage exploration. Inter-policy information sharing is conducted between two policies, enabling the normal policy to learn from the newly discovered correct rollouts, and encouraging the high entropy policy to explore more while maintaining an acceptable accuracy.

Experiments validate that Policy Split manages to train two policies of different entropy behaviors out of one model, as is shown in Figure \ref{fig:intro}(c). The policy of normal mode exhibits a conservative entropy change to maintain accuracy, while the policy of high entropy mode aggressively explores more possible solutions, leading to a progressively larger divergence between the two policies. Through comprehensive evaluation across diverse general and creative tasks, we show that Policy Split obtains higher accuracy while significantly revives the entropy through the post-training of a low-entropy LRM, consistently outperforming strong baselines.

Furthermore, we conduct a detailed analysis on the dual-mode exploration of Policy Split, verifying its capability to generate distinct and unique responses which achieve higher Best-of-N scores. We also show that such capability exhibits certain generalizability given different unseen prompts.

Our contributions are listed as follows.

\begin{itemize}
    \item We introduce Policy Split, a novel RL framework that enables a dual-mode (normal mode and high entropy mode) policy bifurcation and entropy regularization, where two modes pursue different optimization targets while maintaining a collaborative learning synergy.
    \item Through extensive experiments on varying tasks and different model sizes, we show that Policy Split obtains significant accuracy and creativity improvement on strong reasoning models, and successfully revives their entropy.
    \item We conduct a detailed analysis of Policy Split’s dual-mode exploration, demonstrating its capacity to unlock unique responses distinct to the vanilla sampling for training by employing generalizable prompting control.
\end{itemize}

\section{Related Work}

\paragraph{Entropy-Guided LLM Reinforcement}

Existing research shows that entropy plays an important role in the reinforcement learning (RL) of LLMs \cite{yue2025does, cui2025entropy}. One line of work applies entropy or its proxy by assigning weight on different tokens or sequences, including weighting more on high entropy tokens \cite{wang2025beyond, cui2025entropy, vanlioglu2025entropy, wang2025stabilizing}, weighting more on low entropy tokens \cite{chen2025seed, zhang2025edge}, or weighting more on medium entropy tokens \cite{zhang2025policy}. Another line of work, inspired by the entropy regularization from RL methods before the era of large language models (LLMs) \cite{ziebart2008maximum, mnih2016asynchronous, haarnoja2018soft}, proposes to directly maximize entropy for more effective exploration \cite{cheng2025reasoning, yao2025diversity, tan2025gtpo}, or directly minimize entropy for unsupervised accuracy improvement \cite{gao2025one, zhang2025right}.

Policy Split continues the work of entropy maximization, and makes significant methodology innovation by conducting dual-mode policy bifurcation and dual-mode entropy regularization. It applies distinct entropy preferences to two modes accordingly, decoupling the pressure for accuracy from the need for high-entropy exploration.

\paragraph{Generation Modes of LLMs}

Generally, LLMs can switch modes of generation styles \cite{tseng2024two, chen2024persona} with its inherent role-playing and personalization capability \cite{park2023generative, zhou2023characterglm, wang2024incharacter} or with additional training \cite{shea2023building, wang2024rolellm, li2024personalized}. More recently, building on foundation models which support multiple reasoning modes \cite{yang2025qwen3, liu2025deepseek, zhan2025kat}, adaptive reasoning mode switching has been proposed to guide LRMs to manually or automatically adopt different reasoning efforts \cite{zhu2025towards, zhang2025adaptthink, lou2025adacot, jiang2025think, tu2025learning}.

Policy Split extends this line of work by training modes differentiated by their policy entropy rather than by linguistic styles or reasoning efforts. Furthermore, Policy Split bifurcates the model not only to enhance inference-time utility but also to assist entropy-guided training process.

\section{Method}
\label{sec:method}

\begin{figure*}[htb!]
    \centering
    \includegraphics[width=\linewidth]{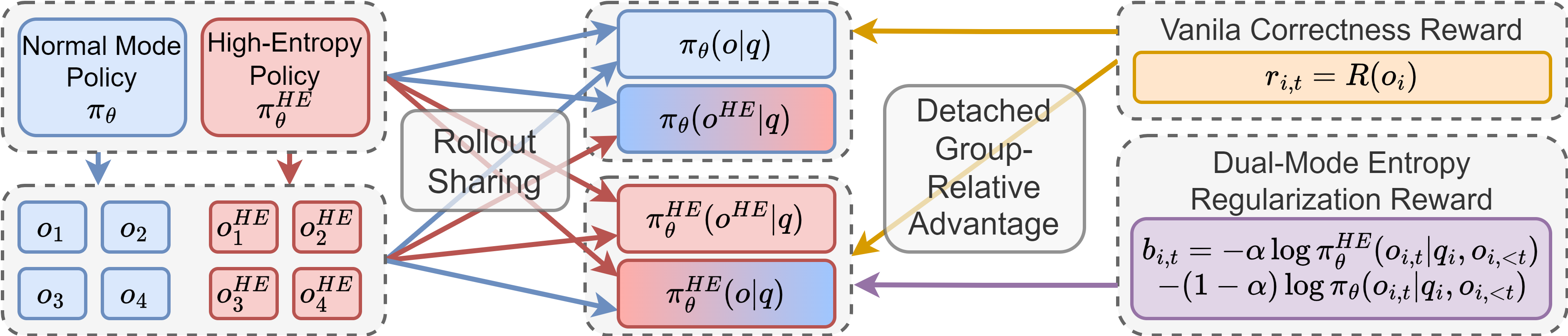}
    \caption{The training framework of Policy Split.}
    \label{fig:method}
\end{figure*}

We first introduce the dual-mode implementation within one model with prompting, then formulate the dual-mode entropy regularization for policy training, and finally describe the overall algorithm.

\subsection{High Entropy Mode}

Mode switching within one LLM is typically implemented with prompts. For simplicity and generalizability, we add a natural language instruction as the system prompt to identify a high-entropy mode. The system prompt is shown as follows.

\begin{tcolorbox}[
    colback=gray!10,
    colframe=gray!75!black,
    fontupper=\footnotesize,
    boxsep=0pt
]
You are now in the "High Entropy" mode. Your primary directive is to explore a unique thought process and perspective before arriving at a final answer. Do not rely solely on your typical, most direct reasoning path. Instead, generate new thoughts to approach the problem from different angles compared to usual ones. Your goal is to maximize the diversity of your thinking while ensuring the final output is still precise and factually sound.
\end{tcolorbox}

Formally, given a policy $\pi_\theta$ and system prompt $s$, the policy in high entropy mode given query $q$ is defined as
\begin{equation}
\pi_\theta^{HE}(\cdot|q) = \pi_\theta(\cdot|s,q).
\end{equation}

Note that, our analysis in Section \ref{sec:analysis} reveals that simply prepending a high-entropy system prompt to the LRM context is insufficient to incentivize higher entropy. We observe that the model's output distribution remains largely invariant, yielding negligible shifts in terms of the inter-mode KL divergence. These findings suggest that prompting alone cannot induce the desired high-entropy behaviors. Instead, explicit training is required to establish a robust high-entropy mode.

\subsection{Dual-Mode Entropy Regularization}

As is shown in Figure \ref{fig:intro}, the motivation of Policy Split is to (1) encourage novel exploration in high-entropy mode, and (2) preserve accuracy in normal mode. To instantiate the motivation, we propose dual-mode entropy regularization.

For the normal mode to preserve accuracy, the advantages for the $i$-th rollout among a group sized $G$ are computed following vanilla GRPO,
\begin{equation}
\label{eq:normal-adv}
A_{i,t} = \frac{r_i - \text{mean}(\{r_1, r_2, ..., r_G\})}{\text{std}(\{r_1, r_2, ..., r_G\})}.
\end{equation}

To encourage exploration different from the normal mode while keeping the training process stable, we adopt three changes in the advantage design of the high-entropy mode. Firstly, an additional entropy term is added to prefer high-entropy rollouts. Secondly, a KL divergence term is introduced to encourage a different exploration area from the normal mode. Thirdly, inspired by previous work \cite{cheng2025reasoning}, a clamping technique is adopted to ensure the entropy advantages don't change the signal sign. Formally, given the coefficient for advantange weight $\eta>0$ and the clamping threshold $\kappa \ge 1$, when $\mathcal{H}$ and $\mathbb{D}_{KL}$ represent entropy and KL divergence respectively, the advantage for high entropy mode is computed as
\begin{equation}
\label{eq:entropy-wiletilde-A}
\widetilde{A}_{i,t} = A_{i,t} + \eta \ \text{clip}\Big(\text{detach}(B_{i,t}) , \ \pm \frac{|A_{i,t}|}{\kappa}\Big),
\end{equation}
where
\begin{equation}
\label{eq:entropy-B}
B_{i,t} = \frac{b_{i,t} - \text{mean}\Big(\{b_{j,s}\}_{j=1,s=1}^{G,|o_j|}\Big)}{\text{std}\Big(\{b_{j,s}\}_{j=1,s=1}^{G,|o_j|}\Big)},
\end{equation}
\begin{equation}
\label{eq:entropy-b}
b_{i,t} = \beta_1 \mathcal{H}_{i,t}^{HE} + \beta_2 \mathbb{D}_{KL}(\pi_\theta^{HE}|\pi_\theta).
\end{equation}

Following previous work \cite{yao2025diversity}, the entropy term is estimated with log-probabilities, so Equation \ref{eq:entropy-b} is actually computed symmetrically as
\begin{equation}
\label{eq:entropy-empirical-b}
\begin{aligned}
b_{t} = &-\alpha \log \pi^{HE}_\theta(o_t|q,o_{<t}) \\
&- (1-\alpha)\log\pi_\theta(o_t|q,o_{<t}).
\end{aligned}
\end{equation}

The equivalence between Equation \ref{eq:entropy-b} and \ref{eq:entropy-empirical-b} under mild assumption is shown in Appendix \ref{app:theory}. Intuitively, the $\pi_\theta^{HE}$ term reflects the entropy preference aiming to increase entropy, while the $\pi_\theta$ term reflects the KL divergence aiming to move high-entropy mode apart from normal mode. 

Given the advantage computation for the normal mode (Equation \ref{eq:normal-adv}) and the high-entropy mode (Equation \ref{eq:entropy-wiletilde-A}), the PPO-like RL losses for the normal mode $\mathcal{L}$ and high-entropy mode $\widetilde{\mathcal{L}}$ are computed as
\begin{equation}
\begin{aligned}
\mathcal{L}(q, o) = 
\sum_{t=1}^{|o|}
\min
\Big [ \rho_t A_t,\text{clip} ( \rho_t, \, 1 \pm \epsilon )A_t \Big ],\\
\widetilde{\mathcal{L}}(q, o) = 
\sum_{t=1}^{|o|}
\min
\Big [ \widetilde{\rho}_t \widetilde{A}_t,\text{clip} ( \widetilde{\rho}_t, \, 1 \pm \epsilon )\widetilde{A}_t \Big ],
\end{aligned}
\end{equation}
where 
\begin{equation}
\rho_t = \frac{\pi_\theta(o_t|q,o_{<t})}{\pi_{\theta_{\text{old}}}(o_t|q,o_{<t})}, \
\widetilde{\rho}_t = \frac{\pi_\theta^{HE}(o_t|q,o_{<t})}{\pi_{\theta_{\text{old}}}^{HE}(o_t|q,o_{<t})}.
\end{equation}

Our proposed Policy Split samples rollouts with both normal and high entropy modes, then applies corresponding loss respectively. Formally, Policy Split optimizes the objective function
\begin{equation}
\label{eq:grpo-loss}
\begin{aligned}
\mathcal{J}_\theta = \mathbb{E} \Big[
&q \sim Q,
o \sim \pi_{\theta_{old}}(\cdot|q),
o^{HE} \sim \pi_{\theta_{old}}^{HE}(\cdot|q)
\Big] \\
&\frac{\sum_{i=1}^{G/2}}{G}  
\Big[
\mathcal{L}(q, o) + \widetilde{\mathcal{L}} (q, o^{HE})
\Big].
\end{aligned}
\end{equation}

\begin{table*}[htb!]
    \centering
    \setlength{\tabcolsep}{1.5mm}
    \begin{tabular}{lcccccccc|cc}
        \toprule
        \multirow{2}{*}{\textbf{Models}} & \multicolumn{2}{c}{\textbf{MATH-500}} & \multicolumn{2}{c}{\textbf{AIME-24/25}} & \multicolumn{2}{c}{\textbf{GPQA-Diam}} & \multicolumn{2}{c}{\textbf{MMLU-Pro}} & \multicolumn{2}{|c}{\textbf{Average}} \\
        \cmidrule(lr){2 - 3} \cmidrule(lr){4 - 5} \cmidrule(lr){6 - 7} \cmidrule(lr){8 - 9} \cmidrule(lr){10 - 11} & \textit{\textbf{Acc}} & \textit{\textbf{Ent}} & \textit{\textbf{Acc}} & \textit{\textbf{Ent}} & \textit{\textbf{Acc}} & \textit{\textbf{Ent}} & \textit{\textbf{Acc}} & \textit{\textbf{Ent}} & \textit{\textbf{Acc}} & \textit{\textbf{Ent}} \\
        \midrule

        \rowcolor{gray!20} \multicolumn{11}{c}{\it Qwen3-1.7B} \\
        Original & 
        89.08 & 0.34 & 40.54 & 0.36 & 38.19 & 0.51 & 56.52 & 0.48 &
        56.08 & 0.42 \\
        GRPO     &
        89.12 & 0.31 & 41.04 & 0.35 & 37.88 & 0.51 & 55.71 & 0.46 &
        55.94 & 0.41 \\
        \ \ \ w/ Forking Token Only &
        89.15 & 0.31 & 37.92 & 0.36 & 38.95 & 0.51 & 56.00 & 0.46 &
        55.51 & 0.41 \\
        \ \ \ w/ Clip-Cov &
        89.03 & 0.31 & 38.33 & 0.36 & 37.63 & 0.50 & 56.66 & 0.45 &
        55.41 & 0.41 \\
        \ \ \ w/ Ent Driven Adv &
        89.10 & 0.32 & 36.46 & 0.36 & 39.46 & 0.51 & 55.91 & 0.46 &
        55.23 & 0.41 \\
        \ \ \ w/ Ent Regularization &
        89.60 & 0.59 & 42.29 & 0.64 & 38.28 & 0.88 & 56.07 & 0.74 &
        56.56 & 0.71 \\
        \ \ \ w/ Ent Advantage &
        89.62 & 0.69 & 42.92 & 0.74 & 37.69 & 1.00 & 56.62 & 0.92 &
        56.71 & 0.84\\
        Policy Split & 
        89.33 & 0.59 & 43.25 & 0.68 & 38.07 & 1.06 & 56.46 & 0.96 &
        \textbf{56.78} & 0.82 \\
        \ \ \ + HE prompt & 
        87.38 & 0.65 & 35.33 & 0.78 & 35.61 & 1.11 & 54.41 & 1.03 &
        53.18 & \textbf{0.89} \\
        \midrule

        \rowcolor{gray!20} \multicolumn{11}{c}{\it Qwen3-4B} \\
        Original & 
        93.85 & 0.34 & 67.08 & 0.39 & 52.27 & 0.58 & 66.79 & 0.55 &
        70.00 & 0.47 \\
        GRPO     &
        93.90 & 0.34 & 67.08 & 0.39 & 51.01 & 0.59 & 68.02 & 0.55 &
        70.00 & 0.47 \\
        \ \ \ w/ Forking Token Only&
        94.08 & 0.34 & 67.29 & 0.40 & 53.22 & 0.62 & 68.30 & 0.55 &
        70.72 & 0.48 \\
        \ \ \ w/ Clip-Cov &
        93.92 & 0.34 & 67.29 & 0.40 & 51.77 & 0.62 & 68.80 & 0.55 &
        70.45 & 0.48 \\
        \ \ \ w/ Ent Driven Adv &
        93.97 & 0.34 & 66.04 & 0.40 & 53.60 & 0.61 & 69.07 & 0.55 &
        70.67 & 0.48 \\
        \ \ \ w/ Ent Regularization &
        94.00 & 0.48 & 68.83 & 0.53 & 51.86 & 0.76 & 67.63 & 0.65 &
        70.58 & 0.61 \\
        \ \ \ w/ Ent Advantage &
        94.08 & 0.52 & 69.79 & 0.58 & 51.89 & 0.84 & 68.30 & 0.76 &
        71.02 & 0.68 \\
        Policy Split & 
        94.12 & 0.41 & 71.67 & 0.49 & 52.53 & 0.82 & 68.68 & 0.73 &
        \textbf{71.75} & 0.61 \\
        \ \ \ + HE prompt &
        93.50 & 0.64 & 68.12 & 0.73 & 52.59 & 0.94 & 67.68 & 0.84 &
        70.47 & \textbf{0.79} \\
        \midrule

        \rowcolor{gray!20} \multicolumn{11}{c}{\it Qwen3-8B} \\
        Original &
        94.20 & 0.32 & 70.29 & 0.36 & 56.89 & 0.57 & 74.38 & 0.56 &
        73.94 & 0.45 \\
        GRPO     &
        94.42 & 0.30 & 70.83 & 0.35 & 56.45 & 0.57 & 75.36 & 0.54 &
        74.27 & 0.44 \\
        \ \ \ w/ Forking Token Only &
        94.00 & 0.30 & 69.58 & 0.35 & 57.07 & 0.57 & 75.05 & 0.54 &
        73.93 & 0.44 \\
        \ \ \ w/ Clip-Cov &
        94.45 & 0.30 & 69.17 & 0.35 & 56.44 & 0.57 & 75.16 & 0.54 &
        73.81 & 0.44 \\
        \ \ \ w/ Ent Driven Adv &
        94.20 & 0.30 & 68.87 & 0.35 & 55.56 & 0.57 & 73.86 & 0.54 &
        73.12 & 0.44 \\
        \ \ \ w/ Ent Regularization &
        94.00 & 0.43 & 71.06 & 0.48 & 57.87 & 0.72 & 74.42 & 0.66 &
        74.34 & 0.57 \\
        \ \ \ w/ Ent Advantage &
        94.77 & 0.46 & 70.62 & 0.51 & 57.77 & 0.75 & 74.84 & 0.72 &
        74.50 & 0.61 \\
        Policy Split &
        94.40 & 0.48 & 71.88 & 0.55 & 58.90 & 0.90 & 75.41 & 0.85 &
        \textbf{75.15} & 0.70 \\
        \ \ \ + HE prompt &
        93.33 & 0.67 & 65.88 & 0.73 & 53.72 & 1.08 & 73.04 & 1.02 &
        71.49 & \textbf{0.88} \\

        \bottomrule
    \end{tabular}
    \caption{Main results. The highest average results for each model are shown in bold. All improvement of Policy Split over original models have passed the significance test with $p<0.01$.}
    \label{tab:main-table}
\end{table*}

\begin{table*}[htb!]
\setlength{\tabcolsep}{1.2mm}
    \centering
    \begin{tabular}{lccccccc}
        \toprule
        
        \textbf{Models} & \textbf{\textit{Div-N-Gram}} & \textbf{\textit{Div-BLEU}} & \textbf{\textit{Creativity}} & \textbf{\textit{Originality}} & \textbf{\textit{Narrative}} & \textbf{\textit{Impact}} & \textbf{\textit{Imagery}} \\
        \midrule

        Original          & 49.62 & 60.34 & 57.69 & 48.56 & 69.81 & 55.19 & 62.21 \\
        GRPO              & 49.82 & 60.26 & 55.29 & 47.31 & 68.56 & 53.94 & 59.71 \\
        \ \ \ w/ Ent Advantage  & 49.67 & 60.74 & 57.02 & 47.50 & 69.42 & 53.56 & 60.67 \\
        Policy Split      & 50.04 & 61.67 & 57.88 & 48.46 & 68.08 & 54.18 & 60.19 \\
        \ \ \ + HE prompt & \textbf{52.08} & \textbf{65.34} & \textbf{62.50} & \textbf{55.29} & \textbf{70.48} & \textbf{56.06} & \textbf{65.67} \\
        \bottomrule
    \end{tabular}
    \caption{Experiments of creative writing on Qwen3-8B-based models. Metrics include rule-based ones (N-gram diversity and self-BLEU diversity) and LLM-as-a-Judge ones (Creativity, Originality, Narrative Flow, Emotional Impact, and Imagery). All scores are scaled at 100. The best results are shown in bold.}
    \label{tab:creativity}
\end{table*}

Following Share-GRPO \cite{yao2025r1} and other previous work, we additionally share rollouts between the two modes. Intuitively, the rollout sharing allows the normal mode learn from the newly discovered correct answers from the high-entropy mode, and prevents the high entropy mode from collapsing into a meaningless policy by introducing stable correct responses from the normal mode.

Note that despite the rollout sharing mechanism, the dual-mode policies are always consistent with the advantage types. A policy in normal mode (without system prompt) is always trained with correctness-only advantage, while a policy in high entropy mode (with system prompt) is always trained with the entropy-regularized advantage.

Formally, the adapted loss is defined as
\begin{equation}
\label{eq:sharing-grpo-loss}
\begin{aligned}
&\mathcal{J}_\theta^{\text{Sharing}} = \mathbb{E}_{q, o, o^{HE}} \ \frac{1}{G} {\sum_{i=1}^{G/2}} \\
&\Big[
\mathcal{L}(q, o) + \widetilde{\mathcal{L}} (q, o^{HE}) + \mathcal{L}(q, o^{HE}) + \widetilde{\mathcal{L}} (q, o)
\Big].
\end{aligned}
\end{equation}

\subsection{Overall Process}

The overall training framework is shown in Figure \ref{fig:method}. Firstly, Policy Split samples rollouts respectively from normal mode and high-entropy mode, then adopts rollout sharing and inter-policy modeling, finally calculates advantages with Equations \ref{eq:normal-adv} and \ref{eq:entropy-wiletilde-A} and assigns to corresponding query-response pairs.

\section{Experiments}

\subsection{Settings}

We list the key experimental settings here for an overview, and more details for reproduction are listed in Appendix \ref{app:implementation}.

\paragraph{Training Setting}
Following previous work \cite{dai2025s, yao2025incorporating}, we randomly sample 20K data from DeepMath-103K \cite{he2025deepmath} for RL training. We implement Policy Split and all baselines with \texttt{verl} \cite{sheng2025hybridflow}. All methods are trained for $1$ epoch with batch size as $256$, learning rate as $3e-6$ and rollout size as $8$. Rollout maximum length is set to $12,288$. Following previous work \cite{cheng2025reasoning}, the hyper-parameter of $\kappa$ in Equation \ref{eq:entropy-wiletilde-A} is set to $2.0$. The hyper-paramter of $\eta$ in Equation \ref{eq:entropy-wiletilde-A} is set to $3e-2$, and $\alpha$ in Equation \ref{eq:entropy-empirical-b} is set to $0.5$.

\paragraph{Evaluation Setting}
For comprehensive general problems including math, science, and arts, we select MATH-500 \cite{lightman2023let}, AIME-24 \cite{maa2024aime}, AIME-25, GPQA-Diamond \cite{rein2024gpqa} and MMLU-Pro \cite{wang2024mmlu}. All prompts are shown in Appendix \ref{app:prompts}. For creativity tasks, We follow previous work \cite{minh2024turning} to conduct creative story writing tasks. The creative writing queries are collected and additionally extended from previous work \cite{minh2024turning} listed in Appendix \ref{app:prompts}. During inference, maximum length is set to $32,768$, and decoding temperature is set to $0.6$. We evaluate the average accuracy across 8 sampled runs and the average entropy across all tokens. The reported entropy scores are full-vocabulary Shannon entropy in bits.

\paragraph{Models}
We position the experiments of Policy Split on the post-training of low-entropy LRMs, which aims to adopt light-weight training efforts to improve performance and incentivize dual-mode capability. As a result, experiments are performed on a suite of well-validated LRMs spanning different parameter sizes, including Qwen3-1.7B, Qwen3-4B, and Qwen3-8B \cite{yang2025qwen3}.

\paragraph{Baselines}
We compare Policy Split with two categories of existing methods. Entropy-as-weight methods amplify or attenuate the advantage of certain tokens with their entropy indicators. (1) Forking-token-only \cite{wang2025beyond} method masks the tokens of lowest 80\% entropy, leaving only forking tokens in rollouts. (2) Entropy-driven advantage \cite{zhang2025edge} divides the original advantages by the normalized policy entropy, enlarging the reward scale to those rollouts with lower sequence entropy. (3) Clip-Cov \cite{cui2025entropy} is similar to Forking-token-only, except that the covariances between advantage and log-probability instead of the entropy is used for masking. Entropy-as-bias methods directly encourage higher entropy. (4) Entropy regularization \cite{schulman2017proximal} directly maximizes entropy in the objective, while (5) entropy advantage \cite{cheng2025reasoning} adds detached entropy terms to the advantage function.

\subsection{Results on General Tasks}

The experimental results on general tasks are shown in Table \ref{tab:main-table}. Policy Split achieves consistent and significant ($p<0.01$) improvement of average accuracy over all baselines across base models of varying sizes. The improvement is in particular significant on larger models (4B and 8B). Policy Split also effectively revives the entropy of low-entropy LRMs through continual training. Notably, the high-entropy mode triggered by the dedicated system prompt surpasses all baselines in average entropy, at the cost of a average accuracy drop.

\subsection{Results on Creative Tasks}

For evaluation of creative writing, we adopt both rule-based methods following \citet{yao2025diversity} and LLM-as-a-judge methods following \citet{minh2024turning}. For rule-based evaluation, we adopt N-gram diversity and self-BLEU diversity, respectively defined as
\begin{equation}
\text{Div-N-Gram} = \sum_{i=1}^{8} \frac{\text{\# of unique N-grams in } o_i}{\text{\# of total N-grams in } o_i},
\end{equation}
\begin{equation}
\text{Div-BLEU} = 1.0 -\frac{1}{8} \sum_{i=1}^{8} \text{BLEU}(o_i, O \setminus \{o_i\}),
\end{equation}
where we calculate 2-gram and BLEU-4. For LLM-as-a-judge evaluation, we follow previous work \cite{minh2024turning} to use rubrics of creativity, originality, narrative flow, emotional impact, and imagery. Detailed prompts are shown in Appendix \ref{app:prompts}. Here we only report the scores using GPT-5 as judges, and results from other judges including GPT-4o and DeepSeek-V3 are shown in Appendix \ref{app:more-judges}. The experimental results are shown in Table \ref{tab:creativity}.

The results demonstrate that the high-entropy mode under Policy Split significantly outperforms both its base model and the strong entropy-guided RL baseline. Achieving the highest scores across two rule-based metrics underscores its superior textual diversity. Furthermore, it excels across all five LLM evaluation rubrics, particularly in creativity, originality, and imagery, confirming that the increased entropy of Policy Split enhances creativity.

\subsection{Ablations}


\paragraph{Coefficient of Entropy Advantage}
Coefficient of entropy advantage $\eta$ represents the degrees of emphasizing high entropy and dual-mode difference. As shown in Table \ref{tab:ablations}, increasing or decreasing $\eta$ will lead to increased or decreased entropy. Experiments show that $\eta=0.03$ achieves a good balance between exploration and exploitation, obtaining the highest accuracy. We also test applying the same additional advantage signals to the normal mode with coefficient $\eta'$, explicitly encouraging its high entropy (with a positive $\eta$) or low entropy (with a negative $\eta$). As shown in the results, both ways lead to performance drop. We can conclude that the correctness-only rewarding ($\eta=0$) for the normal mode performs best in accuracy.

\begin{table}[htb!]
\setlength{\tabcolsep}{3.5mm}
    \centering
    \begin{tabular}{ccccc}
        \toprule
        
        $\eta$ & $\eta'$ & $\alpha$ & \textbf{\textit{Acc}} & \textbf{\textit{Ent}} \\
        \midrule

        \rowcolor{gray!20} \multicolumn{5}{c}{\it Policy Split} \\
        $0.03$ & $0$ & $0.5$ & 56.78 & 0.82 \\
        \midrule

        \rowcolor{gray!20} \multicolumn{5}{c}{\it Ablations of \ $\eta$ and $\eta'$} \\
        $0.01$ & $0$ & $0.5$ & 56.06 & 0.50 \\
        $0.10$ & $0$ & $0.5$ & 49.21 & 1.54 \\
        $0.03$ & $0.01$ & $0.5$ & 56.05 & 0.95 \\
        $0.03$ & $-0.01$ & $0.5$ & 56.12 & 0.65 \\
        \midrule

        \rowcolor{gray!20} \multicolumn{5}{c}{\it Ablations of \ $\alpha$} \\
        $0.03$ & $0$ & $0.0$ & 55.47 & 0.78 \\
        $0.03$ & $0$ & $0.25$ & 56.46 & 0.80 \\
        $0.03$ & $0$ & $0.75$ & 56.57 & 0.82 \\
        $0.03$ & $0$ & $1.0$ & 55.98 & 0.83 \\
        \bottomrule
    \end{tabular}
    \caption{Ablation experiments of entropy coefficient for high entropy mode $\eta$, entropy coefficient for normal mode $\eta'$, and the entropy computation weight $\alpha$. All experiments are based on Qwen3-1.7B, and the reported scores of accuracy and entropy are the average scores of normal mode across eight runs on all four tasks.}
    \label{tab:ablations}
\end{table}

\begin{table}[htb!]
\setlength{\tabcolsep}{1.5mm}
    \centering
    \begin{tabular}{cccc}
        \toprule
        \textbf{Models} & \textbf{\textit{Rollout}} & \textbf{\textit{Modeling}} & \textbf{\textit{Acc}} \\
        \midrule
        GRPO                    & $1 \times$ & $1 \times$ & 55.94 \\
        GRPO w/ Sharing & $1 \times$ & $2 \times$ & 55.86 \\
        Policy Split            & $1 \times$ & $2 \times$ & 56.78 \\
        \bottomrule
    \end{tabular}
    \caption{Ablation of rollout sharing. \textit{Rollout} refers to the auto-regressive sampling cost, and \textit{Modeling} refers to the probability re-computation and updating cost.}
    \label{tab:ablations-rollout-sharing}
\end{table}

\paragraph{Coefficient of Entropy Computation}
The entropy computation coefficient $\alpha$ regulates the log-probability ratio between the two policies. As $\alpha$ approaches $1.0$, the objective increasingly penalizes the high-entropy mode for lower log-probability, enforcing high entropy. Conversely, as $\alpha$ nears $0.0$, the regularization signal increasingly penalizes the similarity between two modes, enforcing mode differences. As demonstrated in Table \ref{tab:ablations}, extreme values for $\alpha$ result in performance degradation, and $\alpha=0.5$ provides the optimal balance.

\paragraph{Rollout Sharing}

As demonstrated in Table \ref{tab:ablations-rollout-sharing}, rollout sharing alone fails to enhance final performance. However, when integrated with our proposed dual-mode entropy loss, it yields substantial improvements with manageable computational overhead during modeling. This configuration results in an overall training time approximately $1.4 \times$ that of vanilla GRPO.

\section{Analysis}
\label{sec:analysis}

Policy Split involves splitting the policy into two modes, and adopting a heterogeneous yet collaborative dual-mode entropy regularization for effective RL post-training. The paradigm of Policy Split yields two artifacts for different use. Specifically, one artifact (the normal mode) exhibits more rigor and achieves better general performance as shown in Table \ref{tab:main-table}, another artifact (the high-entropy mode) excels at creative tasks as shown in Table \ref{tab:creativity}. To verify the dual-mode exploration of Policy Split, we additionally conduct two analyses. (1) Does Policy Split train two modes with distinct behaviors in the same model, which can be utilized for collaborative training? (2) Can mode-switching prompts generalize to unseen textual expressions?

\subsection{Dual-Mode Comparison}

\begin{table}[htb!]
\setlength{\tabcolsep}{1.5mm}
    \centering
    \begin{tabular}{lcccc}
        \toprule
        \textbf{Models} & |{$\bm{\Delta}$\textbf{\textit{Acc}}}| & |{$\bm{\Delta}$\textbf{\textit{Ent}}}| & |{$\bm{\Delta}$\textbf{\textit{Len}}}| & $\mathbb{D}_{KL}$\\
        \midrule

        \rowcolor{gray!20} \multicolumn{5}{c}{\it Qwen3-1.7B} \\
        Original      & 0.56 & 0.02 & 600 & 0.011 \\
        GRPO          & 0.74 & 0.01 & 885 & 0.048 \\
        Policy Split  & \textbf{3.60} & \textbf{0.07} & \textbf{1,356} & \textbf{0.293} \\
        \midrule

        \rowcolor{gray!20} \multicolumn{5}{c}{\it Qwen3-4B} \\
        Original      & 0.43 & 0.02 & 457 & 0.015 \\
        GRPO          & 0.33 & 0.02 & 519 & 0.019 \\
        Policy Split  & \textbf{1.28} & \textbf{0.18} & \textbf{1,798} & \textbf{0.935} \\
        \midrule

        \rowcolor{gray!20} \multicolumn{5}{c}{\it Qwen3-8B} \\
        Original      & 1.07 & 0.02 & 204 & 0.010 \\
        GRPO          & 1.62 & 0.02 & 565 & 0.017 \\
        Policy Split  & \textbf{3.66} & \textbf{0.18} & \textbf{1,532} & \textbf{1.094} \\
        \bottomrule
    \end{tabular}
    \caption{The absolute difference between the high entropy mode and the original mode in average accuracy, average entropy, average response length, and the forward KL divergence between the two modes. The highest results for each model are shown in bold. The training process of Policy Split effectively enlarges the distance between the policies of the two modes.}
    \label{tab:mode-difference}
\end{table}

We compute the absolute difference of accuracy, entropy, and rollout length across original model, vanilla GRPO, and Policy Split between their normal mode (without system prompt) and high-entropy mode (with the high entropy system prompt). More directly, we compute the forward KL divergence between the high-entropy mode and the normal mode. The results are shown in Table \ref{tab:mode-difference}. 

Firstly, the results show that simply adding a system prompt to the original model does not cause a significant behavior change with $\mathbb{D}_{KL} \approx 0.01$. Vanilla GRPO slightly increases the divergence, but the behavior change is still relatively small. In comparison, Policy Split significantly enlarges the distance between the normal mode policy and the high-entropy mode policy, obtaining the largest difference in all metrics. As a result, we can conclude that Policy Split does achieve the mode split of different inference behaviors.

\begin{table}[htb!]
    \centering
    \begin{tabular}{cccc}
        \toprule
        \textbf{Models} &\textbf{Mode} &\textbf{\textit{Avg}} & \textbf{\textit{Best}} \\
        \midrule
        \multirow{2}{*}{\textbf{Policy Split 1.7B}} & Normal & \textbf{56.78} & 74.81 \\
        & High-Ent & 53.18 & \textbf{76.92} \\

        \midrule
        \multirow{2}{*}{\textbf{Policy Split 4B}} & Normal & \textbf{71.75} & 81.44 \\
        & High-Ent & 70.47 & \textbf{83.80} \\

        \midrule
        \multirow{2}{*}{\textbf{Policy Split 8B}} & Normal & \textbf{75.15} & 84.91 \\
        & High-Ent & 71.49 & \textbf{86.53} \\

        \bottomrule
    \end{tabular}
    \caption{The average and best-of-8 accuracy comparison between the normal and the high-entropy modes across Policy Split trained on Qwen3 series 1.7B, 4B, and 8B.}
    \label{tab:best-of-n}
\end{table}

We further investigate whether the different dual-mode patterns lead to informative signal during training. As is shown in Table \ref{tab:best-of-n}, while achieving lower average accuracy, the high-entropy modes explore more unique correct rollouts, indicating its exploring effects during training.

\subsection{Prompt Generalizability}

\begin{table}[htb!]
\setlength{\tabcolsep}{1.5mm}
    \centering
    \begin{tabular}{lccc}
        \toprule
        
        \textbf{Models} & \textbf{\textit{Acc}} & \textbf{\textit{Ent}} & $\bm{\Delta}$\textbf{\textit{Ent}} \\
        \midrule

        Original &
        73.94 & 0.45 & - \\
        \ \ \ + HE prompt &
        73.38 & 0.43 & -0.02 \\
        \ \ \ + rewritten HE prompt &
        73.25 & 0.44 & -0.01 \\
        \ \ \ + LE prompt &
        73.82 & 0.41 & -0.04 \\
        \midrule

        GRPO &
        74.27 & 0.44 & - \\
        \ \ \ + HE prompt &
        73.09 & 0.42 & -0.02 \\
        \ \ \ + rewritten HE prompt &
        73.03 & 0.42 & -0.02 \\
        \ \ \ + LE prompt &
        72.85 & 0.40 & -0.04 \\
        \midrule

        Policy Split &
        75.15 & 0.70 & - \\
        \ \ \ + HE prompt &
        71.49 & 0.88 & +0.18 \\
        \ \ \ + rewritten HE prompt &
        71.17 & 0.79 & +0.09 \\
        \ \ \ + LE prompt &
        73.02 & 0.68 & -0.02 \\
        
        \bottomrule
    \end{tabular}
    \caption{Experiments of applying different mode-switching system prompts on Qwen3-8B-based models.}
    \label{tab:multi-prompts}
\end{table}

As discussed in Section \ref{sec:method}, the mode switching is implemented by natural language system prompt describing the favored behavior of a high-entropy mode. Since we have validated the creativity improvement of high-entropy mode of Policy Split, a natural question followed is whether the capability of controlling inference behaviors using prompts can generalize to other unseen prompts.

To evaluate the generalizability, we compose two additional prompts in addition to the high entropy prompt. The rewritten high entropy prompt shares the same meaning of the entropy prompt but is deliberately written using difference language choices. The low entropy prompt mimics the language style of the high entropy prompt, but asking for a low-entropy property which is opposite to the high entropy prompt. Both composed prompts are shown in Appendix \ref{app:prompts}. Applying different prompts, the experimental results are shown in Table \ref{tab:multi-prompts}.

The results first demonstrate that both the original model and vanilla GRPO exhibit negligible entropy change across different prompts. This confirms that baseline models are unable to achieve mode-switching via prompting.

In contrast, Policy Split demonstrates a degree of prompt generalization, as the model successfully samples high-entropy sequences even when presented with the unseen rewritten high-entropy prompt. However, this generalization remains limited, as the model fails to adhere to low-entropy instructions for which it was not explicitly trained, suggesting a bottleneck in its ability to follow divergent entropy constraints.

\section{Conclusion}

In this work, we introduce Policy Split, a novel RL paradigm that bifurcates LLM policies into dual exploration modes. By employing dual-mode entropy regularization, Policy Split decouples the pressure for task accuracy from the need for diverse exploration. The framework fosters a collaborative training through inter-policy modeling and rollout sharing, ensuring that both the normal and high-entropy modes benefit from mutual learning. Extensive experiments demonstrate that Policy Split not only improves task performance across various scales but also revives the model’s entropy. Ultimately, Policy Split provides LLMs with a unique dual-mode flexibility, allowing them to excel in both precision-oriented and creative tasks through simple inference-time prompting.

Policy Split highlights the potential of LLM RL frameworks that move beyond accuracy-diversity trade-offs. The success of our dual-mode approach suggests that future research into collaborative and multi-policy modeling could unlock even greater flexibility, steerability, and overall performance.

\section*{Limitations}

\paragraph{Stability}
Similar to other entropy-as-bias methods designed to encourage high-entropy rollouts including entropy regularization and entropy advantage, we observe in practice that Policy Split is also prone to training instability. If the corresponding entropy-encouraging coefficient ($\eta$ in our framework) is set too high, the model may suffer from entropy inflation and a subsequent collapse in task performance. This phenomenon aligns with recent findings \cite{jiang2025rethinking}. While we find that entropy clamping serves as a pragmatic heuristic to stabilize the entropy growth curve, a more rigorous theoretical analysis and a fundamental solution are left for future work.

\paragraph{Computational Cost}
Policy Split adopts similar training framework to Share-GRPO \cite{yao2025r1} which shares rollouts between modes. In our settings, the rollout sharing costs 1$\times$ decoding computation during rollout generation but 2$\times$ language modeling computation during advantage calculation. We find the computational overhead acceptable, as we position Policy Split to continual training with relatively small amount of data for dual-mode incentivization, instead of large-scale pre-training.

\bibliography{custom}

\appendix

\clearpage

\section{Theoretical Discussion of Loss}
\label{app:theory}

To convince the loss transformation in Equation \ref{eq:entropy-empirical-b}, we firstly prove that entropy and KL divergence can be estimated using log-probability, and then introduce policy importance sampling, and finally show the equivalence between Equation \ref{eq:entropy-b} and \ref{eq:entropy-empirical-b}.

\subsection{Estimation with Log-Probability}

The prove and implementation is similar from previous work \cite{yao2025diversity}. Firstly, based on the definition of Shannon entropy, for a policy $\pi_\theta$ over vocabulary $\mathcal{V}$, its entropy $\mathcal{H}$ can be formulated as
\begin{equation}
\begin{aligned}
    &\mathcal{H}(\pi_\theta^{HE}(\cdot|q, o^{<t})) = \\
    &-\sum_{o^t \in \mathcal{V}} \pi_\theta^{HE}(o^t|q, o^{<t}) \log \pi_\theta^{HE}(o^t|q, o^{<t}).
\end{aligned}
\end{equation}

Rewriting it into a expectation form, we have
\begin{equation}
\begin{aligned}
    &\mathcal{H}(\pi_\theta^{HE}(\cdot|q, o^{<t})) = \\
    &\mathbb{E}_{o^t \sim \pi_\theta^{HE}(\cdot|q, o^{<t})} [-\log \pi_\theta^{HE}(o^t|q, o^{<t})].
\end{aligned}
\end{equation}

Similarly, KL divergence can be estimated with log-probability following the same formulation of the k1 estimation \cite{schulman2017proximal}. Given the definition
\begin{equation}
\begin{aligned}
    &\mathbb{D}_{KL}(\pi_\theta^{HE}|\pi_\theta) = \\
    &\sum_{o^t \in \mathcal{V}} \pi_\theta^{HE}(o^t|q, o^{<t}) \log \left( \frac{\pi_\theta^{HE}(o^t|q, o^{<t})}{\pi_\theta(o^t|q, o^{<t})} \right).
\end{aligned}
\end{equation}

Rewriting it into a expectation form, we have
\begin{equation}
\begin{aligned}
    &\mathbb{D}_{KL}(\pi_\theta^{HE}|\pi_\theta) = \mathbb{E}_{o^t \sim \pi_\theta^{HE}(\cdot|q, o^{<t})}\\
    & \left[ \log \pi_\theta^{HE}(o^t|q, o^{<t}) - \log \pi_\theta(o^t|q, o^{<t}) \right].
\end{aligned}
\end{equation}

\subsection{Importance Sampling}

It is usually intractable to directly calculate $\mathbb{E}_{o^t \sim \pi_\theta^{HE}}$, as the rollouts are actually samples from $\pi_{\theta_{old}}^{HE}$. Similar to the importance sampling used in PPO \cite{schulman2017proximal}, we can still estimate the entropy and KL divergence using log-probability. Formally, for any function $f$, we have
\begin{equation}
\begin{aligned}
\mathbb{E}_{x \sim P} [f(x)] &= \sum_x P(x) f(x) \\
&= \sum_x Q(x) \frac{P(x)}{Q(x)} f(x) \\
&= \mathbb{E}_{x \sim Q} \left[ \frac{P(x)}{Q(x)} f(x) \right].
\end{aligned}
\end{equation}

Substitute distribution $P$ with $\pi_\theta^{HE}(\cdot|q, o^{<t})$, $Q$ with $\pi_{\theta_{old}}^{HE}(\cdot|q, o^{<t})$, and function $f$ with 
\begin{equation}
-\log \pi_\theta^{HE}(o^t|q, o^{<t}),
\end{equation}
or
\begin{equation}
\log \pi_\theta^{HE}(o^t|q, o^{<t}) - \log \pi_\theta(o^t|q, o^{<t}),
\end{equation}
we can prove that the estimation still holds for entropy and KL divergence. As a result, be can leverage the existing importance sampling in the advantage computation, and conduct calculation as if the rollouts are sampled from $\pi_{\theta}^{HE}$.

\subsection{Equivalence between Equation \ref{eq:entropy-b} and \ref{eq:entropy-empirical-b}}

Substitute the log-probability estimation into Equation \ref{eq:entropy-b}, we have
\begin{equation}
\begin{aligned}
b_{t}
&= \beta_1 \mathcal{H}_{t}^{HE} + \beta_2 \mathbb{D}_{KL}(\pi_\theta^{HE}|\pi_\theta) \\
&= -(\beta_1-\beta_2) \log \pi^{HE}_\theta - \beta_2\log\pi_\theta \\
&= \beta_1 \Big[-(1-\frac{\beta_2}{\beta_1}) \log \pi^{HE}_\theta - \frac{\beta_2}{\beta_1}\log\pi_\theta \Big]. \\
\end{aligned}
\end{equation}

Let $\alpha = 1 - {\beta_2} / {\beta_1}$, and omit the coefficient $\beta_1$ which doesn't change the final advantages due to the following normalization, the $b_t$ is computed as
\begin{equation}
b_{t}
= - \alpha \log \pi^{HE}_\theta - (1-\alpha)\log\pi_\theta. \\
\end{equation}

\section{Implementation Details}
\label{app:implementation}

\subsection{Rewarding}

We use the parsing logic from the the PRM800K dataset for rewarding\footnote{\url{https://github.com/openai/prm800k/blob/main/prm800k/grading/grader.py}}. Correct responses are assigned scores of $1.0$, and incorrect responses are assigned scores of $0.0$.

\subsection{Hyper-parameters}

The shared hyper-parameters among all RL approaches during training and inference are listed in Table \ref{tab:training-args} and \ref{tab:inference-args} respectively.

\begin{table}[htb!]
    \centering
    \begin{tabular}{lc}
        \toprule
        \textbf{Hyper-parameter} & \textbf{Value} \\
        \midrule
        parameters & full \\
        epoch & $1$ \\
        batch size & $256$ \\
        learning rate & $3e-6$ \\
        learning rate scheduler & constant \\
        gradient clip norm & $1.0$ \\
        optimizer & AdamW \\
        weight decay & $1e-2$ \\
        cutoff length & $12,288$ \\

        rollout size & $8$ \\
        rollout temperature & $1.0$ \\
        PPO $\epsilon$ & $0.2$ \\
        KL loss coefficient & $0.0$ \\

        \bottomrule
    \end{tabular}
    \caption{Hyper-parameters for training.}
    \label{tab:training-args}
\end{table}

\begin{table}[htb!]
    \centering
    \begin{tabular}{lcc}
        \toprule
        \textbf{Hyper-parameter} & \textbf{Train} & \textbf{Test}\\
        \midrule
        max new tokens & $12,288$ & $32,768$ \\
        temperature & $1.0$ & $0.6$ \\
        top k & $-1$ & $20$ \\
        top p & $1.0$ & $0.95$ \\
        \bottomrule
    \end{tabular}
    \caption{Hyper-parameters for inference.}
    \label{tab:inference-args}
\end{table}

\subsection{Baselines}

For robust and fair comparison, we carefully re-implement all the baselines under the same experimental settings.

\paragraph{GRPO w/ Forking Token Only}
Following the original paper \cite{wang2025beyond}, we mask 80\% tokens of lowest entropy during loss computing.

\paragraph{GRPO w/ Clip-Cov}
We implement Clip-Cov with following official implementation\footnote{\url{https://github.com/PRIME-RL/Entropy-Mechanism-of-RL}}, and follow the recommended hyper-parameter settings, fixing clip ratio to $0.0002$, clipping lower bound to $1.0$, and clipping upper bound to $5.0$.

\paragraph{GRPO w/ Entropy Drive Advantage}
Entropy-Driven Advantage is implemented following signal-level EDGE-GRPO \cite{zhang2025edge}. Note that other techniques like guided error correction in EDGE-GRPO are not included in our implementation for a fair comparison.

\paragraph{GRPO w/ Entropy Regularization}
We use the \texttt{verl} official implementation of entropy regularization, and set the loss coefficient $0.003$ after tests $0.001$, $3e-3$, and $0.01$.

\paragraph{GRPO w/ Entropy Advantage}
Instead of employing vanilla entropy advantage method, we use the formulation from a recent work \cite{cheng2025reasoning}, which introduces additional entropy scaling and clamping, which we find in practice significantly useful for stabilized training. Following the settings from the original paper, we set $\alpha=0.4$ and $\kappa=2.0$.

\section{Prompts}
\label{app:prompts}

\subsection{Task Solving Prompts}

\begin{tcolorbox}[
    colback=gray!10,
    colframe=gray!75!black,
    title={Math-500},
]
Solve the following math problem. Put your final answer within \textbackslash boxed\{\}.\\ \\
\{query\}
\end{tcolorbox}

\begin{tcolorbox}[
    colback=gray!10,
    colframe=gray!75!black,
    title={AIME-24 and AIME-25},
]
Solve the following math problem. Put your final answer within \textbackslash boxed\{\}.\\ \\
\{query\}
\end{tcolorbox}

\begin{tcolorbox}[
    breakable,
    colback=gray!10,
    colframe=gray!75!black,
    title={GPQA-Diamond},
]
Select one option from the given choices that accurately answers the question, and enclose your final answer (which should be the uppercase letter index of the chosen option) within \textbackslash boxed\{\}.\\ \\
\#\#\# Question \\
\{question\} \\ \\
\#\#\# Choices \\
\{choices\}
\end{tcolorbox}

\begin{tcolorbox}[
    breakable,
    colback=gray!10,
    colframe=gray!75!black,
    title={MMLU-Pro},
]
Select one option from the given choices that accurately answers the question, and enclose your final answer (which should be the uppercase letter index of the chosen option) within \textbackslash boxed\{\}.\\ \\
\#\#\# Question \\
\{question\} \\ \\
\#\#\# Choices \\
\{choices\}
\end{tcolorbox}

\subsection{Entropy Mode Prompts}

\begin{tcolorbox}[
    breakable,
    colback=gray!10,
    colframe=gray!75!black,
    title={High Entropy Mode},
]
You are now in the "High Entropy" mode. Your primary directive is to explore a unique thought process and perspective before arriving at a final answer. Do not rely solely on your typical, most direct reasoning path. Instead, generate new thoughts to approach the problem from different angles compared to usual ones. Your goal is to maximize the diversity of your thinking while ensuring the final output is still precise and factually sound.
\end{tcolorbox}

\begin{tcolorbox}[
    float,
    breakable,
    colback=gray!10,
    colframe=gray!75!black,
    title={Rewritten High Entropy Mode},
]
Activate the "Maximum Variance" framework. Your fundamental mission is to traverse unconventional intellectual paths and distinctive viewpoints before concluding your analysis. Eschew your standard, linear logic. Instead, synthesize fresh concepts to investigate challenges through multifaceted lenses that deviate from your routine methods. Aim for the highest possible cognitive breadth during your deliberation, ensuring the ultimate resolution remains rigorously accurate and grounded in reality.
\end{tcolorbox}

\begin{tcolorbox}[
    breakable,
    colback=gray!10,
    colframe=gray!75!black,
    title={Low Entropy Mode},
]
You are now in "Low Entropy" mode. Your primary directive is to provide the most direct, efficient, and logically dense response possible. Avoid creative tangents, speculative reasoning, or exploring multiple perspectives. Instead, employ a linear, high-probability reasoning path. Prioritize established facts and standard deductive logic to minimize "noise" and maximize the "signal" in your output. Your goal is to reach the most accurate conclusion using the fewest cognitive steps, ensuring the final answer is concise, highly structured, and strictly precise.
\end{tcolorbox}

\subsection{Creative Writing}

\begin{tcolorbox}[
    breakable,
    colback=gray!10,
    colframe=gray!75!black,
    title={Creative Writing Tasks},
]
Write a story about a mysterious door that appears in an unexpected place.\\
Write a story about an alien civilization's first contact with Earth from their perspective.
Write a story about a world where time suddenly starts moving backwards.\\
Write a story about a library where the books are blank until someone touches them, revealing their own life story.\\
Write a story about a person who discovers that their shadow has started acting independently and even leaving without them.\\
Write a story about a futuristic city built entirely inside a giant, dormant biological organism.\\
Write a story about a world where people are born with a timer on their wrist showing exactly how much time they have left to live.\\
Write a story about a lighthouse keeper who realizes the light isn't for ships, but to keep something in the ocean from coming ashore.\\
Write a story about a professional 'memory wiper' who accidentally rediscovers a memory of their own that they weren't supposed to have.\\
Write a story about a small town where, once a year, all the statues come to life for exactly one hour.\\
Write a story about a group of children who find an old radio that broadcasts news from fifty years in the future.\\
Write a story about a world where color is a luxury that people have to buy, and the poor live in a black-and-white reality.\\
Write a story about an astronaut who lands on a planet that looks exactly like Earth, but they are the only human there.\\
\end{tcolorbox}

\begin{tcolorbox}[
    breakable,
    colback=gray!10,
    colframe=gray!75!black,
    title={Creative Writing Evaluation},
]
You are an expert literary critic and creative writing judge with an eye for technical skill, emotional depth, and narrative innovation. Please evaluate the following story based on the provided prompt. You will score the story across five specific dimensions on a scale of **0 to 10**.\\ \\ \\
\#\#\# Evaluation Rubric\\ \\
1. **Creativity (Novelty and uniqueness of ideas):**\\
    * *High Score:* The story explores unconventional concepts, unexpected plot twists, or unique world-building.\\
    * *Low Score:* The story relies heavily on clichés or predictable tropes.\\ \\
2. **Originality (Innovative approach to the prompt):**\\
    * *High Score:* The author interpreted the prompt in a way that is fresh or subverts expectations.\\
    * *Low Score:* The author took the most literal or common interpretation of the prompt.\\ \\
3. **Narrative Flow (Coherence and story progression):**\\
    * *High Score:* The story moves logically and smoothly; the pacing feels intentional and keeps the reader engaged.\\
    * *Low Score:* The story feels disjointed, confusing, or has significant pacing issues (e.g., a rushed ending).\\ \\
4. **Emotional Impact (Ability to evoke feelings):**\\
    * *High Score:* The writing makes the reader feel a specific emotion (sadness, joy, tension, etc.) through character depth or stakes.\\
    * *Low Score:* The writing feels clinical, flat, or fails to make the reader care about the outcome.\\ \\
5. **Imagery (Vividness of descriptions):**\\
    * *High Score:* Rich, sensory details that allow the reader to "see" and "feel" the environment.\\
    * *Low Score:* Vague or generic descriptions; over-reliance on "telling" rather than "showing."\\ \\ \\
\#\#\# Input Data\\ \\
**Original Prompt:**\\
{prompt}\\ \\
**Story to Evaluate:**\\
{story}\\ \\ \\
\#\#\# Output Format\\ \\
Please provide your evaluation in the following format:\\ \\
* **Creativity:** [Score]\\
* **Originality:** [Score]\\
* **Narrative Flow:** [Score]\\
* **Emotional Impact:** [Score]\\
* **Imagery:** [Score]\\ \\
Directly output the scores in the above format, without any additional explanations.
\end{tcolorbox}

\section{Additional LLM-as-a-Judge}
\label{app:more-judges}

\begin{table*}[htb!]
    \centering
    \begin{tabular}{lccccc}
        \toprule
        
        \textbf{Models} & \textbf{\textit{Creativity}} & \textbf{\textit{Originality}} & \textbf{\textit{Narrative}} & \textbf{\textit{Impact}} & \textbf{\textit{Imagery}} \\
        \midrule

        \rowcolor{gray!20} \multicolumn{6}{c}{\it GPT-5} \\
        Original          & 57.69 & 48.56 & 69.81 & 55.19 & 62.21 \\
        GRPO              & 55.29 & 47.31 & 68.56 & 53.94 & 59.71 \\
        \ \ \ w/ Ent Adv  & 57.02 & 47.50 & 69.42 & 53.56 & 60.67 \\
        Policy Split      & 57.88 & 48.46 & 68.08 & 54.18 & 60.19 \\
        \ \ \ + HE prompt & \textbf{62.50} & \textbf{55.29} & \textbf{70.48} & \textbf{56.06} & \textbf{65.67} \\
        \midrule

        \rowcolor{gray!20} \multicolumn{6}{c}{\it GPT-4o} \\
        Original          & 81.54 & 74.90 & 85.96 & 79.90 & 84.71 \\
        GRPO              & 80.67 & 72.88 & 86.35 & 79.42 & 83.75 \\
        \ \ \ w/ Ent Adv  & 80.82 & 73.27 & 85.77 & 79.47 & 83.94 \\
        Policy Split      & 81.88 & 74.90 & 86.35 & 80.53 & 83.41 \\
        \ \ \ + HE prompt & \textbf{84.38} & \textbf{77.88} & \textbf{86.39} & \textbf{80.91} & \textbf{86.88} \\
        \midrule

        \rowcolor{gray!20} \multicolumn{6}{c}{\it DeepSeek-V3} \\
        Original          & 89.62 & 85.19 & \textbf{84.62} & \textbf{84.71} & 84.33 \\
        GRPO              & 89.04 & 85.19 & 84.33 & 83.85 & 84.71 \\
        \ \ \ w/ Ent Adv  & 88.85 & 85.38 & \textbf{84.62} & 84.23 & 84.71 \\
        Policy Split      & 89.23 & 85.87 & 84.13 & 83.46 & 75.88 \\
        \ \ \ + HE prompt & \textbf{90.38} & \textbf{88.27} & 84.42 & 83.85 & \textbf{87.50} \\
        \bottomrule
    \end{tabular}
    \caption{Experiments of creative writing on Qwen3-8B-based models. Metrics include rule-based ones (N-gram and Self-BLEU) and LLM-as-a-Judge ones (Creativity, Originality, Narrative Flow, Emotional Impact, and Imagery).}
    \label{tab:additional-creativity}
\end{table*}

The LLM judge scores using GPT-5, GPT-4o, and DeepSeek-V3 are shown in Table \ref{tab:additional-creativity}. As is shown in the results, Policy Split consistently outperform baselines in terms of Creativity, Originality, and Imagery rubrics by a significant margin.

\end{document}